\documentclass{article}

\usepackage{PRIMEarxiv}

\usepackage[utf8]{inputenc} 
\usepackage[T1]{fontenc}    
\usepackage{hyperref}       
\usepackage{url}            
\usepackage{booktabs}       
\usepackage{amsfonts}       
\usepackage{nicefrac}       
\usepackage{microtype}      
\usepackage{lipsum}
\usepackage{fancyhdr}       
\usepackage{graphicx}       
\usepackage{placeins}
\usepackage{float}
\graphicspath{{media/}}     

\title{Binary Classification for High Dimensional Data using Supervised Non-Parametric Ensemble Method}

\author{
  Nandan Kanvinde, Pinky Gerela \\
  Thakur Institute Of Management Studies, Career Development \& Research \\
  Mumbai 400101, India\\
  \texttt{\{kanvindenandan81,dr.pinkyg5\}@gmail.com} \\
   \And
  Abhishek Gupta, Raunak Joshi \\
  University of Mumbai \\
  Mumbai 400032, India\\
  \texttt{\{abhishek.gupta20001,raunakjoshi.m\}@gmail.com} \\
}

\begin{document}
\maketitle

\begin{abstract}
High dimensional data for classification does create many difficulties for machine learning algorithms. The generalization can be done using ensemble learning methods such as bagging based supervised non-parametric random forest algorithm. In this paper we solve the problem of binary classification for high dimensional data using random forest for polycystic ovary syndrome dataset. We have performed the implementation and provided a detailed visualization of the data for general inference. The training accuracy that we have achieved is 95.6\% and validation accuracy over 91.74\% respectively.
\end{abstract}


\section{Introduction}
Machine Learning \cite{8862451} technique performs predictive analysis and works with different dimensions of data. The high dimensional data is problematic for some of the basic machine learning algorithms. Data can be particularly intended for classification \cite{10.2307/2344237} or regression \cite{Maulud_Abdulazeez_2020} tasks, but high dimensions in the data are independent of the factor. In this paper we try to consider the classification task, especially binary classification \cite{Kumari2017MachineLA} task. Considering the binary classification for high dimensional data we require learning procedure that is not quite basic. Considering a basic linear learning classification algorithm such as Logistic Regression \cite{Cramer2003-ti}, the binary classification is better because the algorithm is intended for it yet after training over high dimensional data with multiple categorical variable, the performance might degrade to a greater extent. The improved algorithms like K-Nearest Neighbors \cite{Guo2003KNNMA}, Support Vector Machines \cite{708428} and CART \cite{1295258} are good in context as compared to Logistic Regression yet have some or the limitation considering the high dimensional data. This is where use of ensemble learning \cite{1999} procedures can prove to be a better solution which is practically an accumulation of weak-set of learners that yield a good result. The main divisions in the area of ensemble learning are bagging \cite{Breiman2004BaggingP} and boosting \cite{Freund1999ASI}. This paper covers the bagging process in detail and boosting is out of context. Now the main task that further requires attention is the data used. Considering the points pertaining to our problem statement, binary classification dataset which high dimensional with categorical variables which will be handled using feature engineering techniques. We contemplated implementing the "Polycystic Ovary Syndrome"\cite{allahbadia2011polycystic} diagnosis dataset. The primary task of the paper is proving that bagging ensemble learning method can prove to be much efficient with high dimensional data and provides a more generalized result as compared to some of the traditional learning procedures.

\section{Implementation}
Bootstrap Aggregation \cite{Lee2019BootstrapAA} is other terminology for bagging \cite{Breiman2004BaggingP} ensemble learning methods. Random Forest \cite{Breiman2004RandomF} is the most commonly used bootstrap aggregation. The process of bootstrap aggregation states segregating the data into pieces with set of rules and later performing an aggregation. The random forest is a supervised non-parametric learning system. Hence it is ensemble learning, it considers the accumulation of weak learners with the bootstrap aggregation system. The basic structure of the random forest uses a set of decision trees \cite{Quinlan2004InductionOD} which are considered as Estimators in the random forest implementation. The estimators when used in large numbers increase the time complexity of the model for learning process. The decision trees by themselves are inefficient as they go into high variance problem when trained on high dimensions. The random forest tackles that problem effectively. The number of decision trees used are influenced by the depth parameter, which indicates the depth of tree starting from the first node. The depth of the random forest can be considered as logarithmic value of the number of estimators. The estimators in random forest for the last layer are considered as 2 times the number of estimators. The random forest is primarily used for generalization \cite{athey2019generalized} of error using out of bag \cite{10.1371/journal.pone.0201904} score as a parameter. The samples which are not trained nor tested are used by the out of bag score for checking the efficiency of data. Considering the parameters used in the implementation using scikit-learn \cite{scikit-learn,sklearn_api} for random forest, the maximum depth of the forest used is 8, estimators which are also known as decision trees are 100, minimal sample split is 23 and minimum sample leaves are 2 respectively.

\section{Results}
\subsection{Analysis of Data}

\begin{figure}[h!]
    \centering
    \includegraphics[scale=0.6]{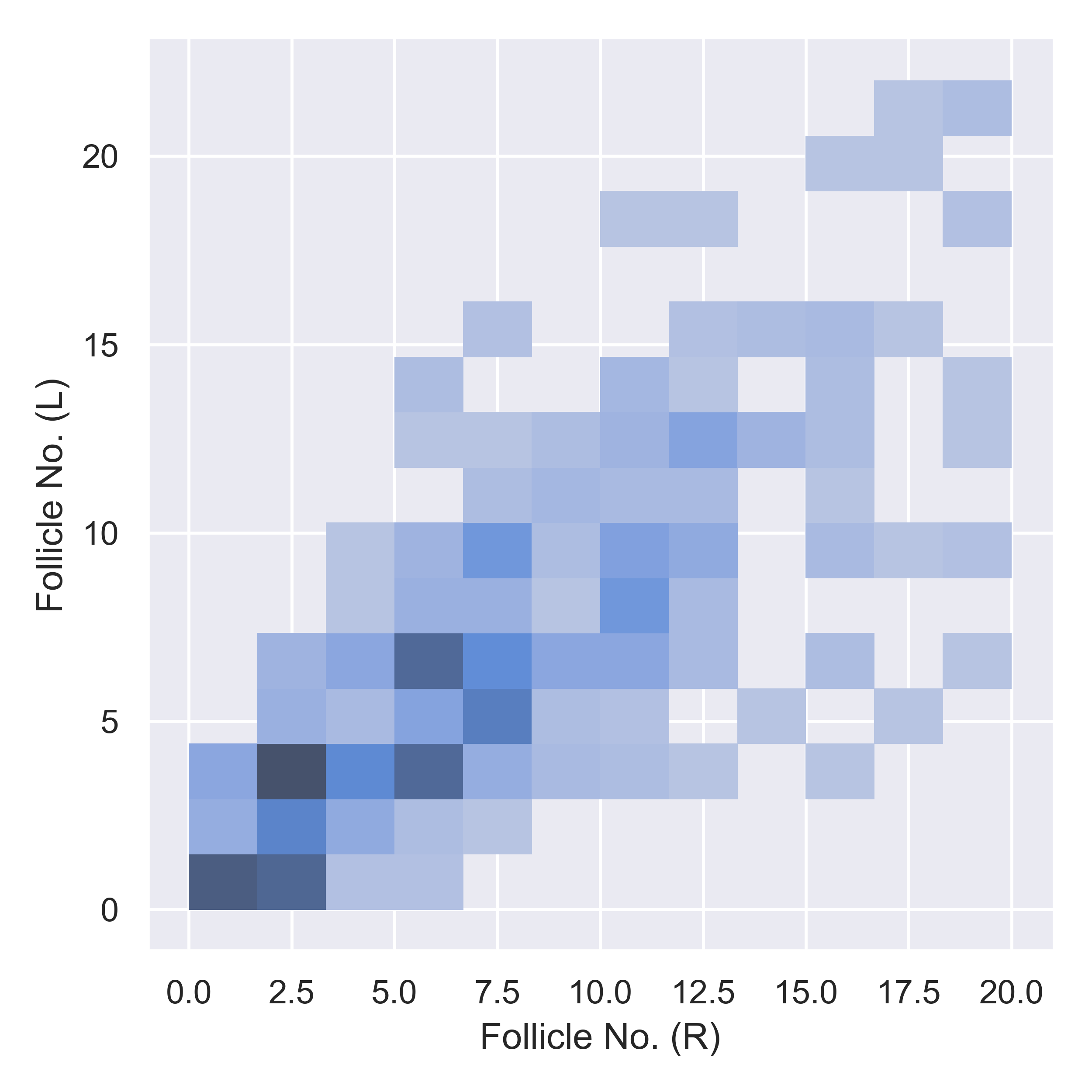}
    \caption{Follicles Correlation}
    \label{fig:figa}
\end{figure}

\begin{figure}[h!]
    \centering
    \includegraphics[scale=0.55]{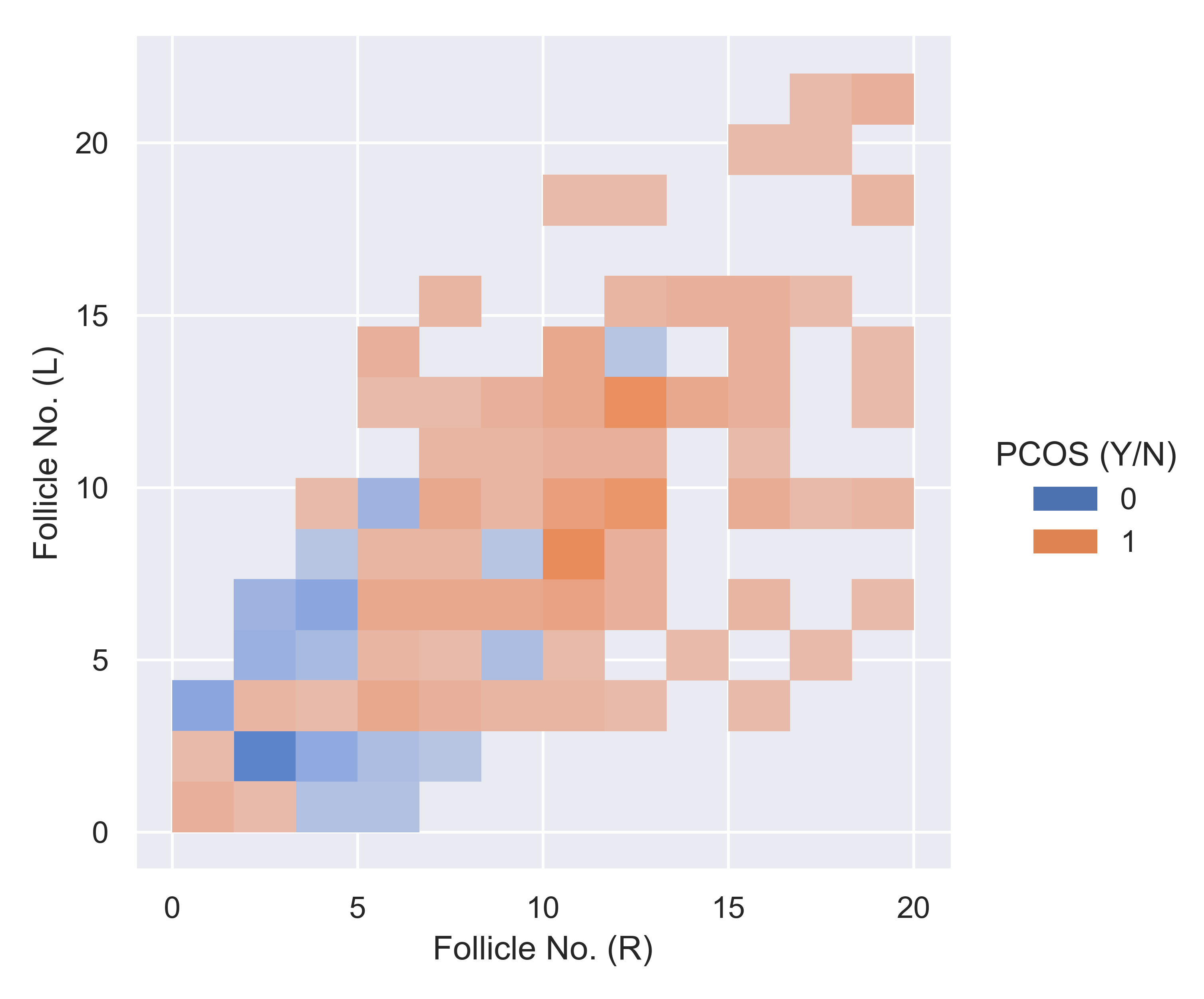}
    \caption{Follicles affected Correlation}
    \label{fig:figb}
\end{figure}

Many implementations of the PCOS is done using boosting methods \cite{gupta2021succinct}, discriminant analysis \cite{gupta2022discriminant}, stacked generalizations \cite{nair2022combining} and deep learning \cite{gupta2022residual}, but analysis of the data is done most precisely in this paper. Analysis of the data in varied processes that can derive the inferences is important. Visualizations\cite{sadiku2016data} are necessary because they can point out very subtle points in the dataset. The figure \ref{fig:figa} visualizes the follicles with its correlation where the correlation is considered from left to right where darker the color, higher the correlation. The follicles that are affected by PCOS are given in figure \ref{fig:figb} where the affected PCOS 
values can be depicted as orange in the figure whereas blue for not affected.

\begin{figure}[htbp]
\centerline{\includegraphics[scale=0.4]{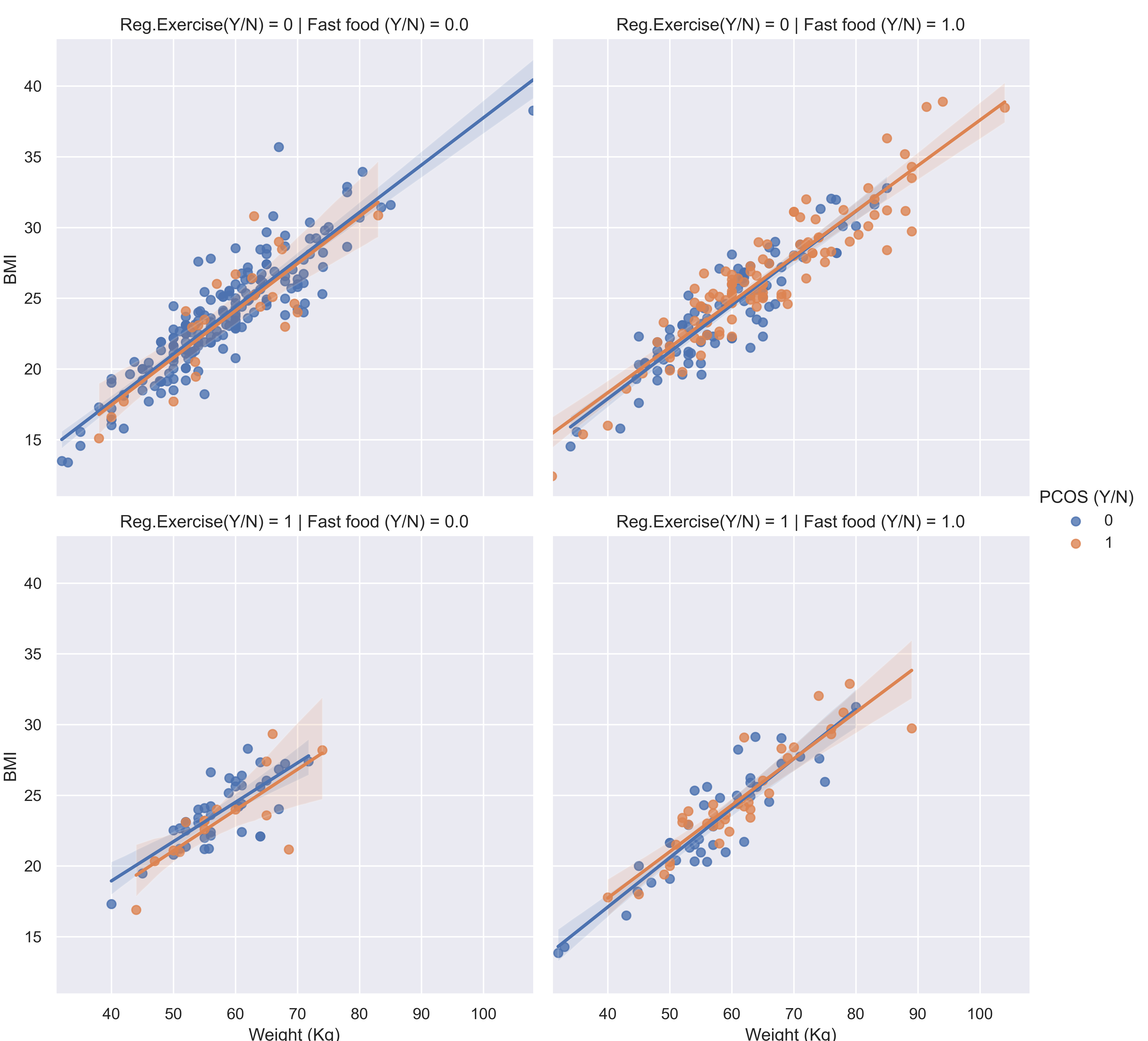}}
\caption{PCOS Exercise Effects}
\label{fig:c}
\end{figure}

The physical activity has influence over the affected PCOS. The figure \ref{fig:c} depicts affects of PCOS when examined with physical activity and junk food consumption. The section where the utilization of junk food is good and regular physical activity is negative, PCOS does affect the highest.

\FloatBarrier

\begin{figure}[htbp]
\centerline{\includegraphics[scale=0.7]{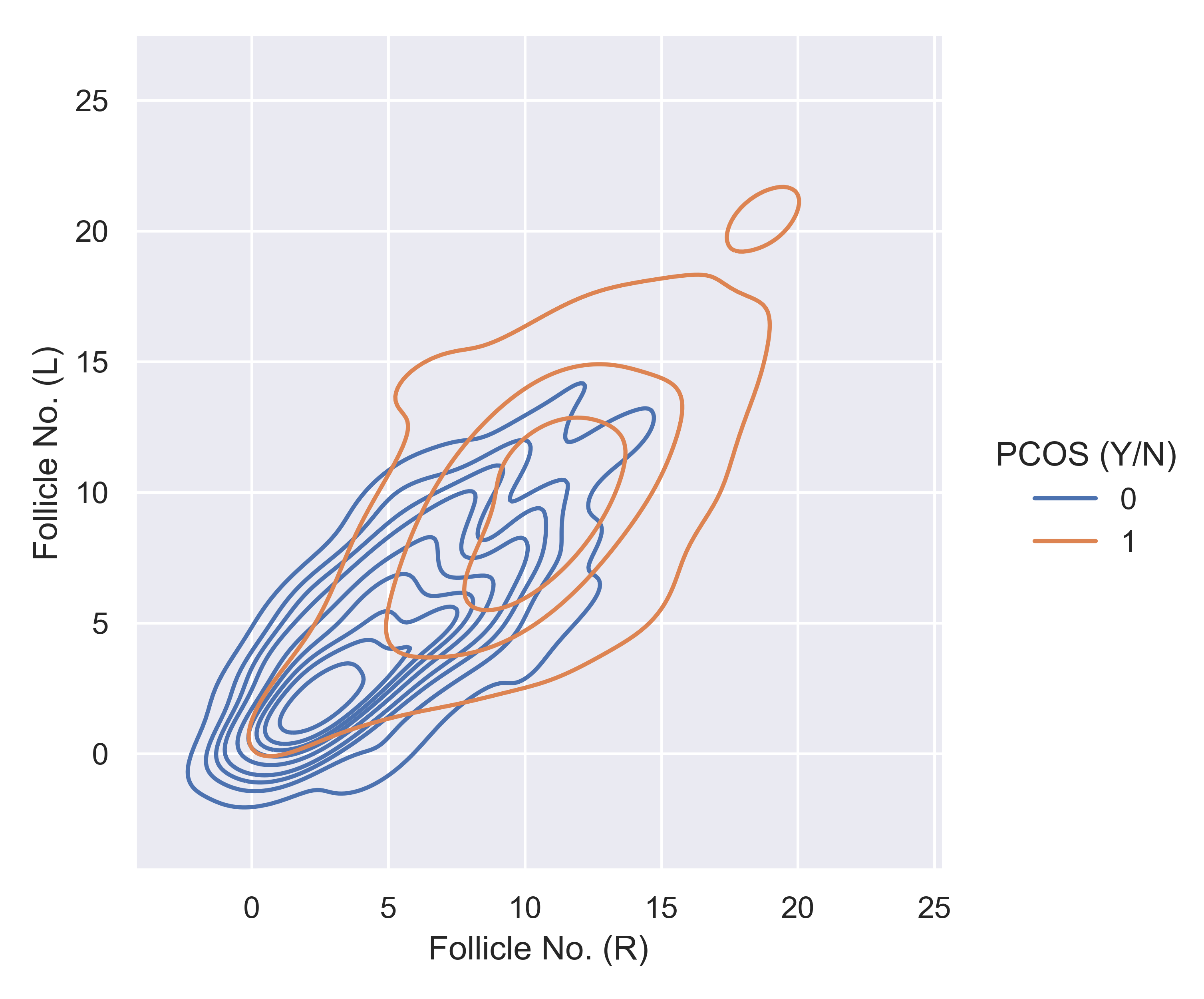}}
\caption{Follicles KDE}
\label{fig:g}
\end{figure}

The figure \ref{fig:g} is depiction of a Bi-variate KDE \cite{chen2017tutorial} Graph. It provides one the continuous PDF curve in specific dimensions for distributions. It is comparatively simple for interpreting than comparing along line scatters. This could be further applied for giving graphs with respect to the classes.

\FloatBarrier

\begin{figure}[htbp]
\centerline{\includegraphics[scale=0.6]{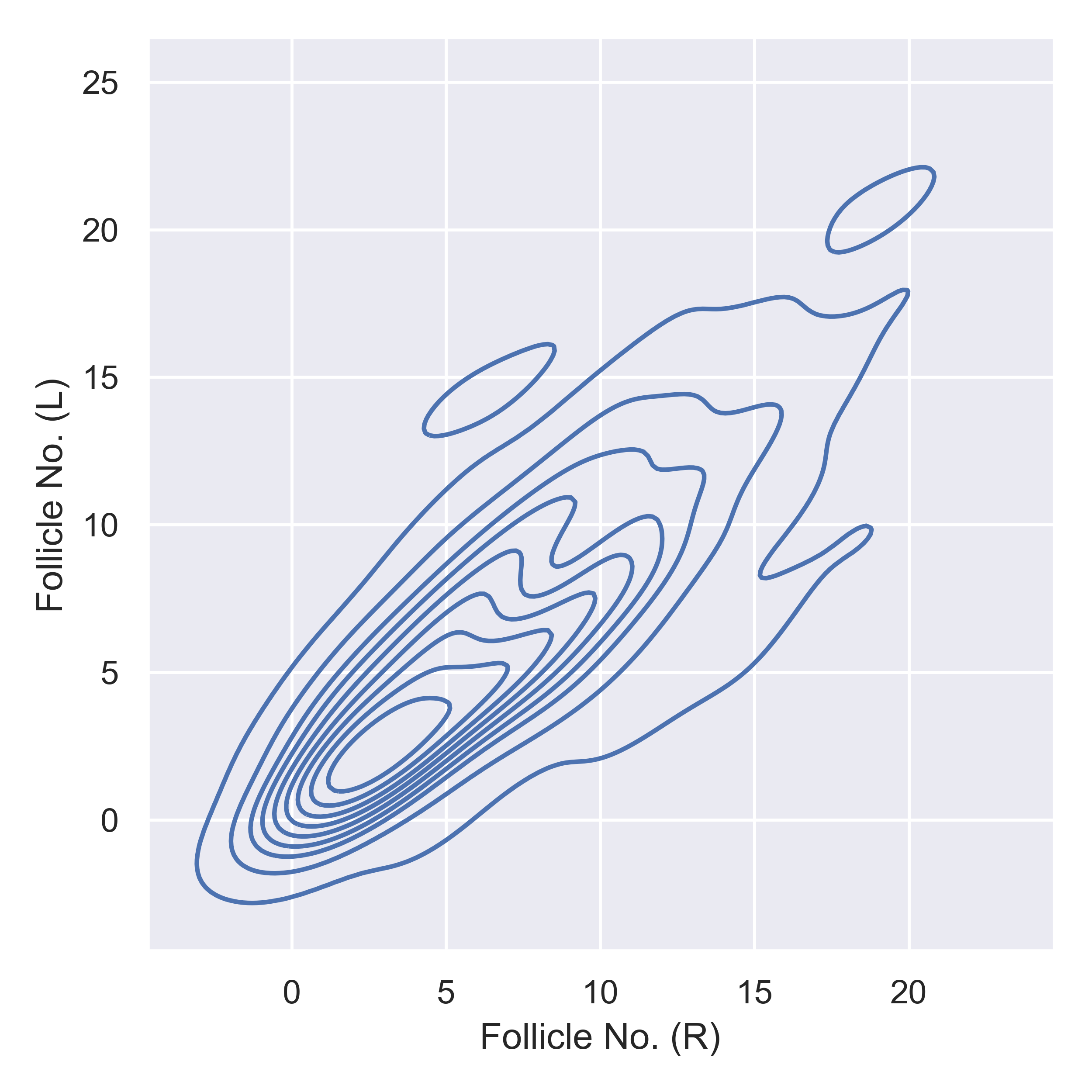}}
\caption{Labeled Follicle KDE}
\label{fig:h}
\end{figure}

The figure \ref{fig:h} provides the Bi-variate KDE graph for follicles in accordance to the classes. The understandability is perceivable as one can certainly infer the affected range of PCOS under the distribution.

\FloatBarrier

\subsection{Precision and Recall}

The Precision and Recall \cite{powers2020evaluation} are basic metrics for all classification problems. Confusion Matrix \cite{Ting2017} focuses on the rudimentary elements like true positives, false positives, true negatives and false negatives for p and r abbreviations for precision and recall. The figure \ref{fig:h} provides the Bi-variate KDE graph for follicles in accordance to the classes. The understandability is perceivable as one can certainly infer the affected range of PCOS under the distribution.

\begin{table}[htbp]
\centering
\caption{Macro and Weighted Averages for P and R}\label{tab:a}
\begin{tabular}{lll}
\toprule
Types & Precision & Recall \\
\midrule
Macro & 0.93 & 0.87 \\
Weighted & 0.91 & 0.91 \\
\bottomrule
\end{tabular}
\end{table}

Precision and Recall have 2 different types known as macro and weighted average. The macro considers all the individual classes into consideration with the unweighted average. 

\begin{table}[htbp]
\centering
\caption{Individual Labels for Precision and Recall}\label{tab:b}
\begin{tabular}{lll}
\toprule
Types & Precision & Recall \\
\midrule
Class 0 & 0.89 & 0.99 \\
Class 1 & 0.96 & 0.75 \\
\bottomrule
\end{tabular}
\end{table}

\subsection{F-Score}

F-Score \cite{Sokolova2006BeyondAF} is the precision and recall harmonic average. Even F-Score has macro and weight averages. The table \ref{tab:c} provides precise depiction of the F-Score.

\begin{table}[H]
\centering
\caption{F-Score for Macro and Weight Average}\label{tab:c}
\begin{tabular}{lll}
\toprule
Metric & Macro & Weight \\
\midrule
F-Score & 0.89 & 0.90 \\
\bottomrule
\end{tabular}
\end{table}

\begin{table}[H]
\centering
\caption{Individual Labels F-Score}\label{tab:d}
\begin{tabular}{lll}
\toprule
Metric & Class 0 & Class 1 \\
\midrule
F-Score & 0.94 & 0.84 \\
\bottomrule
\end{tabular}
\end{table}

Same as different classes observations were available in table \ref{tab:b}, depiction for separate classes are present for F-Score. The table \ref{tab:d} does provide precise F-Score with respect to all the single classes. Separate classes depict the results observations on individual levels. Combination does make a difference in long run.

\subsection{Receiver Operating Characteristic}
\begin{figure}[ht]
\centerline{\includegraphics[scale=0.55]{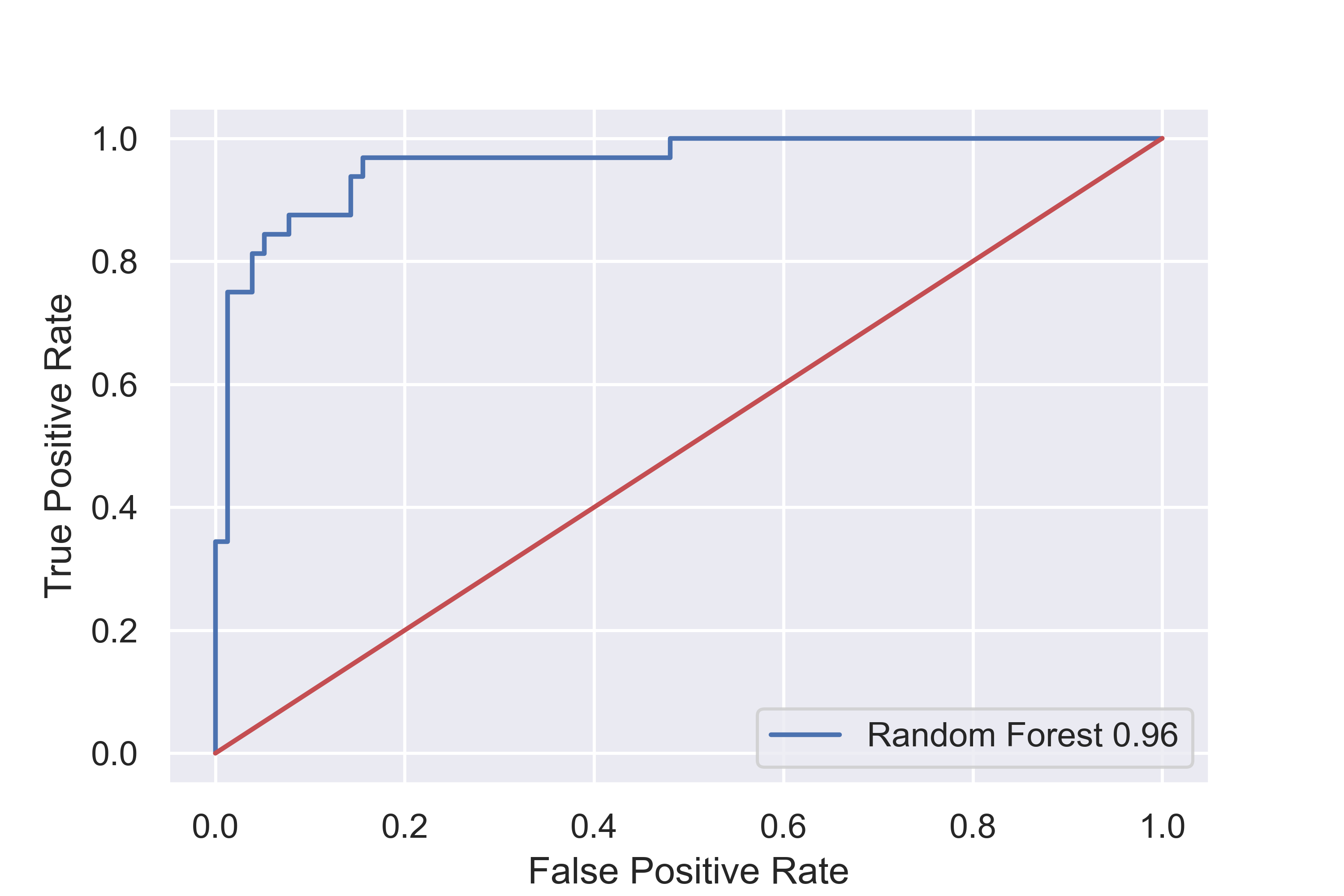}}
\caption{RoC Curve and AuC}
\label{fig:f}
\end{figure}
\FloatBarrier
The Receiver Operating Characteristic \cite{Fawcett2006AnIT} abbreviated as RoC Curve is graphical depiction for binary classification only that provides the algorithm, quality for surpassing specific threshold values. The area that falls above the threshold is considered as Area Under Curve \cite{Bradley1997TheUO}. Values closer to 100\% are considered to be good and closer to 0\% are considered weakly performing. The basic elements that form the RoC Curve are true positive rate \cite{Wang2013} and false positive rate which are basically the sensitivity and specificity. The figure \ref{fig:f} shows the random forest performed with RoC curve. The solid scarlet color line that bypasses through middle is threshold which once crossed indicates a good metric for consideration of RoC. The random forest in this case performs good and gives the score 98\% area under curve.

\section{Conclusion}
If we consider the binary classification problem with high dimensional data the use of bagging ensemble method proves to be effective at generalizing the model and also covering the drawbacks of traditional algorithms. This paper covers a perfect implementation of the random forest for polycystic ovary syndrome dataset. The paper gives a detailed visualization based analysis of the data along with varied metrics that cover the reach of the algorithm. This paper will definitely enlighten many researchers on subtle topics and their implementations which we are proud to be a part of it.

\bibliographystyle{unsrt}  
\bibliography{references}

\begin{thebibliography}{10}

\bibitem{8862451}
Susmita Ray.
\newblock A quick review of machine learning algorithms.
\newblock In {\em 2019 International Conference on Machine Learning, Big Data,
  Cloud and Parallel Computing (COMITCon)}, pages 35--39, 2019.

\bibitem{10.2307/2344237}
R.~M. Cormack.
\newblock A review of classification.
\newblock {\em Journal of the Royal Statistical Society. Series A (General)},
  134(3):321--367, 1971.

\bibitem{Maulud_Abdulazeez_2020}
Dastan Maulud and Adnan~M. Abdulazeez.
\newblock A review on linear regression comprehensive in machine learning.
\newblock {\em Journal of Applied Science and Technology Trends},
  1(4):140--147, Dec. 2020.

\bibitem{Kumari2017MachineLA}
Roshan Kumari and Saurabh~Kr. Srivastava.
\newblock Machine learning: A review on binary classification.
\newblock {\em International Journal of Computer Applications}, 160:11--15,
  2017.

\bibitem{Cramer2003-ti}
J~S Cramer.
\newblock The origins of logistic regression.
\newblock {\em SSRN Electron. J.}, 2003.

\bibitem{Guo2003KNNMA}
Gongde Guo, Hui Wang, David~A. Bell, Yaxin Bi, and Kieran R.~C. Greer.
\newblock Knn model-based approach in classification.
\newblock In {\em OTM}, 2003.

\bibitem{708428}
M.A. Hearst, S.T. Dumais, E.~Osuna, J.~Platt, and B.~Scholkopf.
\newblock Support vector machines.
\newblock {\em IEEE Intelligent Systems and their Applications}, 13(4):18--28,
  1998.

\bibitem{1295258}
H.R. Bittencourt and R.T. Clarke.
\newblock Use of classification and regression trees (cart) to classify
  remotely-sensed digital images.
\newblock In {\em IGARSS 2003. 2003 IEEE International Geoscience and Remote
  Sensing Symposium. Proceedings (IEEE Cat. No.03CH37477)}, volume~6, pages
  3751--3753 vol.6, 2003.

\bibitem{1999}
D.~Opitz and R.~Maclin.
\newblock Popular ensemble methods: An empirical study.
\newblock {\em Journal of Artificial Intelligence Research}, 11:169–198, Aug
  1999.

\bibitem{Breiman2004BaggingP}
Leo Breiman.
\newblock Bagging predictors.
\newblock {\em Machine Learning}, 24:123--140, 2004.

\bibitem{Freund1999ASI}
Yoav Freund and Robert~E. Schapire.
\newblock A short introduction to boosting.
\newblock 1999.

\bibitem{allahbadia2011polycystic}
Gautam~N Allahbadia and Rubina Merchant.
\newblock Polycystic ovary syndrome and impact on health.
\newblock {\em Middle East Fertility Society Journal}, 16(1):19--37, 2011.

\bibitem{Lee2019BootstrapAA}
Tae-Hwy Lee, Aman Ullah, and Ran Wang.
\newblock Bootstrap aggregating and random forest.
\newblock 2019.

\bibitem{Breiman2004RandomF}
Leo Breiman.
\newblock Random forests.
\newblock {\em Machine Learning}, 45:5--32, 2004.

\bibitem{Quinlan2004InductionOD}
J.~Ross Quinlan.
\newblock Induction of decision trees.
\newblock {\em Machine Learning}, 1:81--106, 2004.

\bibitem{athey2019generalized}
Susan Athey, Julie Tibshirani, and Stefan Wager.
\newblock Generalized random forests.
\newblock {\em The Annals of Statistics}, 47(2):1148--1178, 2019.

\bibitem{10.1371/journal.pone.0201904}
Silke Janitza and Roman Hornung.
\newblock On the overestimation of random forest’s out-of-bag error.
\newblock {\em PLOS ONE}, 13(8):1--31, 08 2018.

\bibitem{scikit-learn}
F.~Pedregosa, G.~Varoquaux, A.~Gramfort, V.~Michel, B.~Thirion, O.~Grisel,
  M.~Blondel, P.~Prettenhofer, R.~Weiss, V.~Dubourg, J.~Vanderplas, A.~Passos,
  D.~Cournapeau, M.~Brucher, M.~Perrot, and E.~Duchesnay.
\newblock Scikit-learn: Machine learning in {P}ython.
\newblock {\em Journal of Machine Learning Research}, 12:2825--2830, 2011.

\bibitem{sklearn_api}
Lars Buitinck, Gilles Louppe, Mathieu Blondel, Fabian Pedregosa, Andreas
  Mueller, Olivier Grisel, Vlad Niculae, Peter Prettenhofer, Alexandre
  Gramfort, Jaques Grobler, Robert Layton, Jake VanderPlas, Arnaud Joly, Brian
  Holt, and Ga{\"{e}}l Varoquaux.
\newblock {API} design for machine learning software: experiences from the
  scikit-learn project.
\newblock In {\em ECML PKDD Workshop: Languages for Data Mining and Machine
  Learning}, pages 108--122, 2013.

\bibitem{gupta2021succinct}
Abhishek~M Gupta, Sannidhi~S Shetty, Raunak~M Joshi, and Ronald~Melwin Laban.
\newblock Succinct differentiation of disparate boosting ensemble learning
  methods for prognostication of polycystic ovary syndrome diagnosis.
\newblock In {\em 2021 International Conference on Advances in Computing,
  Communication, and Control (ICAC3)}, pages 1--5. IEEE, 2021.

\bibitem{gupta2022discriminant}
Abhishek Gupta, Himanshu Soni, Raunak Joshi, and Ronald~Melwin Laban.
\newblock Discriminant analysis in contrasting dimensions for polycystic ovary
  syndrome prognostication.
\newblock {\em arXiv preprint arXiv:2201.03029}, 2022.

\bibitem{nair2022combining}
Sruthi Nair, Abhishek Gupta, Raunak Joshi, and Vidya Chitre.
\newblock Combining varied learners for binary classification using stacked
  generalization.
\newblock {\em arXiv preprint arXiv:2202.08910}, 2022.

\bibitem{gupta2022residual}
Abhishek Gupta, Sruthi Nair, Raunak Joshi, and Vidya Chitre.
\newblock Residual-concatenate neural network with deep regularization layers
  for binary classification.
\newblock In {\em 2022 6th International Conference on Intelligent Computing
  and Control Systems (ICICCS)}, pages 1018--1022. IEEE, 2022.

\bibitem{sadiku2016data}
Matthew Sadiku, Adebowale~E Shadare, Sarhan~M Musa, and Cajetan~M Akujuobi.
\newblock Data visualization.

\bibitem{chen2017tutorial}
Yen-Chi Chen.
\newblock A tutorial on kernel density estimation and recent advances, 2017.

\bibitem{powers2020evaluation}
David M.~W. Powers.
\newblock Evaluation: from precision, recall and f-measure to roc,
  informedness, markedness and correlation, 2020.

\bibitem{Ting2017}
Kai~Ming Ting.
\newblock {\em Confusion Matrix}, pages 260--260.
\newblock Springer US, Boston, MA, 2017.

\bibitem{Sokolova2006BeyondAF}
Marina Sokolova, Nathalie Japkowicz, and Stanisaw Szpakowicz.
\newblock Beyond accuracy, f-score and roc: A family of discriminant measures
  for performance evaluation.
\newblock In {\em Australian Conference on Artificial Intelligence}, 2006.

\bibitem{Fawcett2006AnIT}
Tom Fawcett.
\newblock An introduction to roc analysis.
\newblock {\em Pattern Recognit. Lett.}, 27:861--874, 2006.

\bibitem{Bradley1997TheUO}
Andrew~P. Bradley.
\newblock The use of the area under the roc curve in the evaluation of machine
  learning algorithms.
\newblock {\em Pattern Recognit.}, 30:1145--1159, 1997.

\bibitem{Wang2013}
Haiying Wang and Huiru Zheng.
\newblock {\em True Positive Rate}, pages 2302--2303.
\newblock Springer New York, New York, NY, 2013.

\end{thebibliography}

\end{document}